\documentclass[letterpaper]{article} 
\usepackage{aaai25}  
\usepackage{times}  
\usepackage{helvet}  
\usepackage{courier}  
\usepackage[hyphens]{url}  
\usepackage{graphicx} 
\def\UrlFont{\rm}  
\usepackage{natbib}  
\usepackage{caption} 
\usepackage{algorithm}
\usepackage{algorithmic}

\usepackage{amsmath} 
\usepackage{booktabs}
\usepackage[table,xcdraw]{xcolor}
\usepackage{newfloat}
\usepackage{listings}
\DeclareCaptionStyle{ruled}{labelfont=normalfont,labelsep=colon,strut=off} 
\floatstyle{ruled}
\newfloat{listing}{tb}{lst}{}
\floatname{listing}{Listing}
\title{A Comprehensive Evaluation of the Sensitivity of Density-Ratio Estimation Based Fairness Measurement in Regression}

\author {
    Abdalwahab Almajed\textsuperscript{\rm 1},
    Maryam Tabar\textsuperscript{\rm 2},
    Peyman Najafirad\textsuperscript{\rm 2}
}
\affiliations {
    \textsuperscript{\rm 1}Imam Abdulrahman Bin Faisal University, Dammam, Saudi Arabia\\
    \textsuperscript{\rm 2}The University of Texas at San Antonio, San Antonio, Texas, USA\\
    analmajed@iau.edu.sa, maryam.tabar@utsa.edu, peyman.najafirad@utsa.edu
}

\usepackage{bibentry}

\begin{document}

\maketitle

\begin{abstract}
  The prevalence of algorithmic bias in Machine Learning (ML)-driven approaches has inspired growing research on measuring and mitigating bias in the ML domain. Accordingly, prior research studied how to measure fairness in regression which is a complex problem. 
In particular, recent research proposed to formulate it as a density-ratio estimation problem and relied on a Logistic Regression-driven probabilistic classifier-based approach to solve it. However, there are several other methods to estimate a density ratio, and to the best of our knowledge, prior work did not study the sensitivity of such fairness measurement methods to the choice of underlying density ratio estimation algorithm. To fill this gap, this paper develops a set of fairness measurement methods with various density-ratio estimation cores and thoroughly investigates how different cores would affect the achieved level of fairness. Our experimental results show that the choice of density-ratio estimation core could significantly affect the outcome of fairness measurement method, and even, generate inconsistent results with respect to the relative fairness of various algorithms. These observations suggest major issues with density-ratio estimation based fairness measurement in regression and a need for further research to enhance their reliability.
\end{abstract}

%

\section{Introduction}

ML-driven systems have been incorporated into several areas (such as health care and financial sector) that directly impacts different aspects of individual's lives \cite{1-doi:10.1126/science.aap8062}. However, ML models, in some cases, exhibit biases in their outcomes, which could inherent from various sources such as data \cite{3-mehrabi2021survey}. For example, an ML model (COMPAS), which predicts whether an offender would commit another crime was found to be biased against African-Americans compared to Caucasian offenders, which in turn, could affect this group's chances of being released \cite{27-propublica_compas_analysis}. 
With the growth in ML applications and the possibility of biases in their outcome, effective measurement and mitigation of such bias is more essential than ever \cite{2-whitehouse_ai_executive_order_2023}.

To effectively handle biases in these models, one must first investigate whether a bias exists, and measure the level of bias.
Fairness metrics are easy to measure in classification, as it can be assessed using standard metrics derived from a model's confusion matrix \cite{29-10.1145/3194770.3194776}. In contrast, measuring fairness in regression is complex and computationally challenging \cite{7-steinberg2020fairness}. The reason for this complexity is that regression problems involves continuous target variables, leading to significant variability in outcomes when the target values change continuously \cite{6-calders2013controlling}.

Past literature has introduced various methods to approximate fairness in regression \cite{9-agarwal2019fair,5-berk2017convex,10-shah2022selective,7-steinberg2020fairness}. In particular, recent work \cite{7-steinberg2020fairness,8-baniecki2021dalex} proposed to use a density ratio  approximation approach to estimate fairness in  regression. 
However, this study primarily focused on using a single density ratio method for the approximation task, leaving uncertainty about the sensitivity of the results to different underlying density ratio estimation models, beyond Logistic Regression-driven Probabilistic Classification-based density ratio estimation. 
To address this gap, we aim to explore the sensitivity of density ratio estimation-based fairness measurement algorithms by examining how they are affected by the choice of the underlying probabilistic classifier (in the Probabilistic Classification-based method) and the choice of underlying density-ratio estimation approach.

To this end, first, we focus on the Probabilistic Classification-based density-ratio estimation approach to fairness measurement \cite{7-steinberg2020fairness}. In particular, we investigate the impact of using alternative probabilistic classifiers, such as Lasso, Ridge \cite{26-scikit-learn,22-sugiyama2012least}, and  Kernel Logistic Regression \cite{31-Kernel_Logistic_Regression}, and compare their results with those of the original Logistic Regression classifier to study how these alternative models affect the returned fairness outcome. The experimental results reveal a significant inconsistency in approximated fairness values in certain circumstances. In particular, while the use of Logistic Regression, Ridge, and Lasso results in a high correlation between the measured fairness values (and achieved Spearman correlations of consistently above 0.85), the outcome of these methods are not highly correlated with that of models relying on kernel-based probabilistic classifiers. For instance, the correlation between the output of Kernel Logistic Regression-based method with Polynomial kernel and Ridge-based method for measuring the Independence metric are (0.15, -0.35, 0.13) across three regression problems, which in turn, suggest a poor alignment between the fairness values estimated by these methods. These show high sensitivity of the Probabilistic Classification-based fairness measurement methods to the choice of underlying probabilistic classifier.

Next, inspired by these observations, we expand our study to investigate how the returned fairness measures could change if other density-ratio estimation approaches (other than the Probabilistic Classification-based one) are used as the core estimation algorithm for measuring fairness.  
To investigate this matter, we utilize methods such as Least Squares Importance Fitting (LSIF) and Unconstrained Least Squares Importance Fitting (uLSIF) \cite{18-kanamori2009least,35-makiyama_densratio} that are built on the ratio matching approach \cite{16-sugiyama2010density}. As a result, similar patterns emerge when using this group of density-ratio estimation methods. In particular, the Spearman correlation between the output of methods using LSIF and uLSIF ($\alpha$ = 0.75) for measuring Sufficiency are (0.1, 0.37, -0.19). Additionally, their output does not necessarily align with the Probabilistic Classification-based approach; for example, the correlation show values of (-0.45*, -0.45*, -0.12)  between LSIF and Logistic Regression across three regression problems.
These observed inconsistencies show high sensitivity of density-ratio based fairness measurement methods. 

In addition, we investigate potential reasons behind such inconsistencies in the outcomes of Probabilistic Classification-based approaches. To this end, we generate 40 synthetic datasets with varying data distributions for privileged and unprivileged groups, and then, assess the fairness outcomes returned by various fairness measurement methods. As a result, we observed that the level of inconsistency among fairness measurement methods increases as the gap between the two groups increases, which can occur due to the high probabilities (e.g., values higher than 0.999) predicted by classifiers in case of large gaps. While the high predicted probabilities along with gap between data distributions of privileged and unprivileged groups could be one potential reason, our analysis suggests that factors beyond these might also contribute to the observed inconsistencies.

Such a sensitivity is a major problem as it can lead to conflicting conclusions about the fairness of ML models. 
For instance, consider a model used in the healthcare sector to diagnose patients with cancer \cite{36-dankwa2022artificial}. While the model may appear fair based on one fairness assessment approach, it could still exhibit bias against certain group in practice. Thus, we strongly urge the development of more robust and dependable methods for quantifying fairness in regression, particularly across various conditions and data distributions.

\section{Literature Review}
The need to ensure that ML models operate in a fair and equitable manner has grown as their use in decision-making has expanded in recent years. To enforce fairness in ML models, we must first detect and measure the bias in their outcomes. There are three fundamental notions of fairness commonly used to measure bias: Independence, Separation, and Sufficiency \cite{12-caton2024fairness,7-steinberg2020fairness}. Independence measures the extent to which the model's prediction is dependent on a sensitive feature \cite{7-steinberg2020fairness}. Separation extends Independence by considering the actual value \cite{7-steinberg2020fairness}. Sufficiency, on the other hand, evaluates the extent to which the actual value is conditionally independent of the sensitive feature given the model’s outcome. \cite{7-steinberg2020fairness}. Their mathematical representations can be shown as follows: assuming that $A$, $Y_{true}$, and $Y_{pred}$ refer to the sensitive feature, ground-truth value, and a model's prediction (respectively), Independence, Separation, and Sufficiency were defined as $Y_{pred} \perp \!\!\! \perp A$, $Y_{pred} \perp \!\!\! \perp A | Y_{true}$, and $Y_{true} \perp \!\!\! \perp A | Y_{pred}$, respectively \cite{7-steinberg2020fairness}.

 Unfortunately, the computation of these metrics in regression (i.e., when  $Y_{pred}$ and $Y_{true}$ are continuous) is highly challenging, and intractable, as opposed to the classification domain. 
To simplify the computation of fairness in regression problems and overcome the difficulties of having continuous target variables, 
one approach proposed applying discretization to convert the continuous variable into a categorical variable with many categories \cite{9-agarwal2019fair}.
Another method involved comparing the model's output for pairs of similar instances across different groups, adapted from fairness concepts in classification \cite{5-berk2017convex}. Furthermore, previous research proposed a Kolmogorov-Smirnov distance based method to evaluate fairness \cite{14-chzhen2020fair}.  Also, a novel fairness concept called Monotonic Selective Risk \cite{10-shah2022selective} has been introduced for selective regression problems, where the model can abstain from making a prediction if its confidence is low. This approach measured the risk for each subgroup as coverage is reduced, ensuring that the reduction does not result in an increase in mean squared error for any subgroup \cite{10-shah2022selective}. In particular, recent research proposed a Probabilistic Classification-based density-ratio estimation approach to measure fairness in regression \cite{7-steinberg2020fairness}.

However, Almajed et al. \shortcite{15-almajed2024consistency} discovered that these fairness measurement methods for a specific fairness metric do not necessarily exhibit consistent outcome regarding the relative fairness of various ML models. To be more specific, they developed several ML models, then measure the fairness of their predictions in terms of a specific fairness metric using various approximation algorithms, and finally, evaluated the correlation between the outcome of each pair of approximation methods. As a result, they found that the density ratio estimation-based approach \cite{7-steinberg2020fairness} frequently demonstrated poor alignment and a higher level of inconsistency with all other fairness estimation methods in regression. Unfortunately, those studies \cite{7-steinberg2020fairness,15-almajed2024consistency} have several major limitations: (1) The exploration of density ratio approaches has been limited to probabilistic classification, leaving other promising techniques unexplored in this context, or (2) They only relied on Logistic Regression as the core classifier,  limiting our understanding of the method's full capabilities and sensitivities. 

To address these limitations, this study aims to comprehensively investigate the sensitivity of the density ratio estimation-based fairness measurement method to the choice of underlying classifiers and density ratio estimation approaches. Primarily, this paper studies the sensitivity of this approach by utilizing different probabilistic classifiers, such as Ridge, Lasso \cite{26-scikit-learn,22-sugiyama2012least} and Kernel Logistic Regression \cite{31-Kernel_Logistic_Regression}, 
along with other density ratio estimation methods, such as LSIF and uLSIF \cite{18-kanamori2009least}.
By testing these different approaches, this study aims to provide a more clear understanding of the behavior of density ratio estimation-based fairness measurement method under various conditions, which lead to a better understanding of its strengths or limitations.

\section{Problem Formulation}

In this section, we provide a formal definition of the Probabilistic Classification-based approach for fairness measurement in regression. According to \cite{7-steinberg2020fairness}, the Independence, Separation and Sufficiency fairness metrics can be formulated as
\( \frac{P(Y_{\text{pred}} \mid A = 1)}{P(Y_{\text{pred}} \mid A = 0)} \), 
 \( \frac{P(Y_{\text{pred}} \mid Y_{\text{true}}, A = 1)}{P(Y_{\text{pred}} \mid Y_{\text{true}}, A = 0)} \), and  
\( \frac{P(Y_{\text{true}} \mid Y_{\text{pred}}, A = 1)}{P(Y_{\text{true}} \mid Y_{\text{pred}}, A = 0)} \) density ratios, respectively.
In theory, there are several ways to approximate such density ratios \cite{30-Sugiyama_Suzuki_Kanamori_2012}. One way is to use Probabilistic Classification-based estimation method, similar to \cite{7-steinberg2020fairness}. In fact, that method first applies the Bayes rule on all probability functions in nominators and denominators (e.g., $\frac{P(Y_{pred} \mid A = 1)}{P(Y_{pred} \mid A = 0)}$ would become equivalent to $\frac{P(A = 1 \mid Y_{pred}) \cdot P(A = 0)}{P(A = 0 \mid Y_{pred}) \cdot P(A = 1)}$), and then, uses a probabilistic classifier (such as Logistic Regression) to estimate $P(A = 1 \mid Y_{pred})$, and accordingly $P(A = 0 \mid Y_{pred})$ \cite{7-steinberg2020fairness}. 

However, there are several probabilistic classifiers that can be used to approximate $P(A = 1 \mid Y_{pred})$, and in this study, we investigate how the choice of classifier could impact the outcome of  fairness measurement methods.
In addition to the aforementioned Probabilistic Classification-based approach, there are other density-ratio density estimation approaches, such as ratio matching \cite{16-sugiyama2010density}, that can approximate original ratios directly from the available samples, and this paper studies how different choices for density ratio estimation would impact the outcome of fairness measurement methods.

\section{Datasets}
In this study, we mainly utilize three publicly available datasets that are chosen for their relevance and frequent use in fairness-related studies, such as \cite{15-almajed2024consistency}.
\begin{enumerate}
    \item The Law School dataset \cite{32-law_school_wightman1998lsac}: This dataset, containing around 20,000 records, includes information about law school admissions, GPAs, and personal demographic details. The task is to predict the student's GPA.
    \item The Communities and Crime dataset \cite{33-misc_communities_and_crime_183}: This dataset, containing around 2,000 records, provides socio-economic and demographic information about various communities along with crime levels. The goal is predict the percentage of crime in the each community.
    \item The Insurance dataset \cite{34-lantz2019machine}: This dataset, which contains around 1,300 records, includes personal and health-related information, and costs of medical charges. The goal is to predict the cost of insurance for a patient.
\end{enumerate}

\section{Sensitivity of Fairness Measurements to the Choice of Probabilistic Classifier }

In this paper, first, we explore the impact of probabilistic classifier on the output of the Probabilistic Classification-driven density ratio estimation based approaches. In fact, prior work \cite{7-steinberg2020fairness, 8-baniecki2021dalex} relies on only one probabilistic classifier, such as Logistic Regression, as the underlying classifier in their framework for estimating fairness metrics in regression. However, as previously noted, the sensitivity of this approach to the choice of classifier has not been thoroughly examined. This lack of testing raises concerns about the reliability and robustness of the approximation approach, and whether the use of alternative classifiers could lead to varying interpretations of the fairness of ML models.

\begin{table*}[]
\caption{Spearman correlations between the output of pairs of fairness measurement methods with various underlying probabilistic classifiers when measuring Independence, Separation and Sufficiency (shown in each column/row). Results are presented across Law School, Communities and Crime, and Insurance datasets, respectively.}
\label{tab:indep_q1_updated}
\resizebox{\textwidth}{!}{%
\begin{tabular}{lllllll}
\hline
Fairness Metric &  & Logistic Regression & Ridge & Lasso & KLR\_Gaussian & KLR\_Polynomial \\ \hline
Independence & Logistic Regression & - & (1.00*, 1.00*, 1.00*) & (1.00*, 0.93*, 0.94*) & (0.84*, 0.11, -0.49*) & (0.15, -0.35, 0.13 \\
 & Ridge & (1.00*, 1.00*, 1.00*) & - & (1.00*, 0.93*, 0.94*) & (0.84*, 0.11, -0.49*) & (0.15, -0.35, 0.13) \\
 & Lasso & (1.00*, 0.93*, 0.94*) & (1.00*, 0.93*, 0.94*) & - & (0.85*, 0.02, -0.4) & (0.15, -0.31, 0.19) \\
 & KLR\_Gaussian & (0.84*, 0.11, -0.49*) & (0.84*, 0.11, -0.49*) & (0.85*, 0.02, -0.40) & - & (0.10, -0.58*, -0.67*) \\
 & KLR\_Polynomial & (0.15, -0.35, 0.13) & (0.15, -0.35, 0.13) & (0.15, -0.31, 0.19) & (0.1, -0.58*, -0.67*) & - \\ \hline
 
Separation & Logistic Regression & - & (1.00*, 1.00*, 1.00*) & (1.00*, 0.94*, 0.93*) & (0.89*, 0.12, -0.4) & (0.39, -0.07, -0.02) \\
 & Ridge & (1.00*, 1.00*, 1.00*) & - & (1.00*, 0.94*, 0.93*) & (0.89*, 0.12, -0.4) & (0.39, -0.07, -0.02) \\
 & Lasso & (1.00*, 0.94*, 0.93*) & (1.00*, 0.94*, 0.93*) & - & (0.89*, 0.05, -0.34) & (0.39, -0.1, 0.0) \\
 & KLR\_Gaussian & (0.89*, 0.12, -0.4) & (0.89*, 0.12, -0.4) & (0.89*, 0.05, -0.34) & - & (0.34, 0.59*, -0.59*) \\
 & KLR\_Polynomial & (0.39, -0.07, -0.02) & (0.39, -0.07, -0.02) & (0.39, -0.1, 0.0) & (0.34, 0.59*, -0.59*) & - \\ \hline

 Sufficiency & Logistic Regression & - & (1.00*, 1.00*, 1.00*) & (0.99*, 1.00*, 0.87*) & (0.60*, 0.60*, 0.07) & (-0.36, -0.19, -0.37) \\
 & Ridge & (1.00*, 1.00*, 1.00*) & - & (0.99*, 1.00*, 0.87*) & (0.60*, 0.60*, 0.07) & (-0.36, -0.19, -0.37) \\
 & Lasso & (0.99*, 1.00*, 0.87*) & (0.99*, 1.00*, 0.87*) & - & (0.59*, 0.60*, -0.01) & (-0.35, -0.19, -0.32) \\
 & KLR\_Gaussian & (0.60*, 0.60*, 0.07) & (0.60*, 0.60*, 0.07) & (0.59*, 0.60*, -0.01) & - & (-0.66*, -0.42, -0.70*) \\
 & KLR\_Polynomial & (-0.36, -0.19, -0.37) & (-0.36, -0.19, -0.37) & (-0.35, -0.19, -0.32) & (-0.66*, -0.42, -0.70*) & - \\ \hline
\end{tabular}%
}
\end{table*}

\subsection{Methodology}
To address this gap, our study replaces the Logistic Regression core \cite{7-steinberg2020fairness, 8-baniecki2021dalex} with a range of alternative probabilistic classifiers, including Ridge, Lasso \cite{24-wang2013tikhonov,26-scikit-learn}, and Kernel Logistic Regression \cite{31-Kernel_Ridge_Classifier}. Then, it follows the same methodology as in \cite{15-almajed2024consistency} to maintain consistency with existing literature, and here, we provide a summary for completeness. The study employs 22 diverse ML models, ranging from traditional algorithms to neural networks applied to three distinct datasets (namely, Law School \cite{32-law_school_wightman1998lsac}, Communities and Crime \cite{33-misc_communities_and_crime_183}, and Insurance \cite{34-lantz2019machine} datasets) separately, and then, the prediction of those ML models serves as an input of fairness measurement methods. This approach enables a thorough analysis of fairness metrics across various methods and contexts.

We then conduct a comparative analysis of the results, assessing the sensitivity of the fairness measurement methods across each pair of methods with different core classifiers to identify any variations in the method's behavior. This comparison helps us understand whether the use of various classifiers could lead to significant changes in fairness outcomes and contributes to better understanding the sensitivity of fairness estimation approaches. To be more specific, we use Spearman correlation to assess the relationships between the output of each pair of fairness measurement methods (we also computed Kendall correlation, with the corresponding results provided in the appendix)\footnote{Statistically significant with p-value of 0.05 are indicated with an asterisk (*).}. Additionally, we utilize plots to visually represent the correlations between different measurement methods for similar fairness metrics. Our code base is available at \url{https://github.com/wahab1412/Fairness_measurement}.

\subsection{Experimental Results}
Table \ref{tab:indep_q1_updated} presents the obtained Spearman correlation values for the fairness metrics of Independence, Separation, and Sufficiency. In particular, it shows the Spearman correlations (across three datasets) between the outcome of each pair of fairness measurement methods that employ different classifiers within the Probabilistic Classification-based density ratio estimation framework. The analysis of Spearman correlation values shows a mixed result regarding the correlation of obtained fairness values. In fact, the use of non-kernel based classifiers (such as Logistic Regression, Lasso, and Ridge) consistently results in strong correlations among the estimated fairness values, with Spearman ranging from 0.87 to 1 across all metrics and datasets. 
However, Ridge and Kernel Logistic Regression (with Polynomial kernel) show Spearman values of (-0.36, -0.19, -0.37), when being used to measure the Sufficiency metric. Similarly, for measuring Independence, the Spearman correlations between the output of models relying on kernel-based Logistic Regression are (0.1, -0.58*, -0.67). These observations suggest high sensitivity of Probabilistic Classification-based density ratio estimation fairness measurement methods to the choice of underlying classifier.

\begin{table*}[t]
\caption{Spearman correlations between the output of pairs of fairness measurement methods with various density ratio estimation approaches when measuring Independence, Separation and Sufficiency (shown in each column/row). Results are presented across Law School, Communities and Crime, and Insurance datasets, respectively.
}

\label{dens_results_sep}
\resizebox{\textwidth}{!}{%
\begin{tabular}{llrrrrrrr}
\toprule
Corr. Metric & Fairness Measurement & \multicolumn{1}{l}{Logistic Regression} & \multicolumn{1}{l}{KLR\_Polynomial} & \multicolumn{1}{l}{LSIF} & \multicolumn{1}{l}{uLSIF $\alpha$ = 0.25} & \multicolumn{1}{l}{uLSIF $\alpha$ = 0.5} & \multicolumn{1}{l}{uLSIF $\alpha$ = 0.75} & \multicolumn{1}{l}{uLSIF $\alpha$ = 1} \\ \midrule
Independence & Logistic Regression & - & (0.04, -0.65*, 0.23) & (0.51*, 0.28, 0.08) & (0.63*, 0.17, -0.12) & (0.52*, 0.04, -0.37) & (0.32, 0.33, 0.33) & (-0.44*, 0.33, -0.0) \\
 & KLR\_Polynomial & (0.04, -0.65*, 0.23) & - & (0.09, -0.26, 0.54*) & (0.04, -0.36, 0.31) & (0.34, -0.06, 0.28) & (0.36, -0.33, 0.36) & (0.05, -0.33, 0.5*) \\
 & LSIF & (0.51*, 0.28, 0.08) & (0.09, -0.26, 0.54*) & - & (0.76*, 0.26, 0.02) & (0.46*, -0.02, -0.02) & (0.26, 0.37, 0.05) & (-0.28, 0.37, 0.08) \\
 & uLSIF $\alpha$ = 0.25 & (0.63*, 0.17, -0.12) & (0.04, -0.36, 0.31) & (0.76*, 0.26, 0.02) & - & (0.77*, 0.29, 0.51*) & (0.3, 0.72*, 0.35) & (-0.18, 0.72*, 0.27) \\
 & uLSIF $\alpha$ = 0.5 & (0.52*, 0.04, -0.37) & (0.34, -0.06, 0.28) & (0.46*, -0.02, -0.02) & (0.77*, 0.29, 0.51*) & - & (0.45*, 0.49*, 0.39) & (-0.02, 0.49*, 0.42) \\
 & uLSIF $\alpha$ = 0.75 & (0.32, 0.33, 0.33) & (0.36, -0.33, 0.36) & (0.26, 0.37, 0.05) & (0.3, 0.72*, 0.35) & (0.45*, 0.49*, 0.39) & - & (-0.18, 1.0*, 0.72*) \\
 & uLSIF $\alpha$ = 1 & (-0.44*, 0.33, -0.0) & (0.05, -0.33, 0.5*) & (-0.28, 0.37, 0.08) & (-0.18, 0.72*, 0.27) & (-0.02, 0.49*, 0.42) & (-0.18, 1.0*, 0.72*) & - 
 \\ \midrule
 Separation & Logistic Regression & - & (0.28, 0.69*, -0.01) & (0.82*, nan, 0.4) & (0.61*, nan, -0.07) & (0.39, nan, -0.35) & (0.34, nan, 0.33) & (0.1, nan, nan) \\
 & KLR\_Polynomial & (0.28, 0.69*, -0.01) & - & (0.03, nan, 0.3) & (0.53*, nan, 0.49*) & (0.48*, nan, 0.02) & (0.5*, nan, -0.33) & (0.07, nan, nan) \\
 & LSIF & (0.82*, nan, 0.4) & (0.03, nan, 0.3) & - & (0.36, nan, 0.19) & (0.31, nan, 0.04) & (0.14, nan, 0.33) & (-0.21, nan, nan) \\
 & uLSIF $\alpha$ = 0.25 & (0.61*, nan, -0.07) & (0.53*, nan, 0.49*) & (0.36, nan, 0.19) & - & (0.7*, nan, 0.33) & (0.53*, nan, 0.12) & (0.1, nan, nan) \\
 & uLSIF $\alpha$ = 0.5 & (0.39, nan, -0.35) & (0.48*, nan, 0.02) & (0.31, nan, 0.04) & (0.7*, nan, 0.33) & - & (0.3, nan, -0.2) & (-0.06, nan, nan) \\
 & uLSIF $\alpha$ = 0.75 & (0.34, nan, 0.33) & (0.5*, nan, -0.33) & (0.14, nan, 0.33) & (0.53*, nan, 0.12) & (0.3, nan, -0.2) & - & (0.15, nan, nan) \\
 & uLSIF $\alpha$ = 1 & (0.1, nan, nan) & (0.07, nan, nan) & (-0.21, nan, nan) & (0.1, nan, nan) & (-0.06, nan, nan) & (0.15, nan, nan) & -  \\ \midrule
 Sufficiency & Logistic Regression & - & (-0.38, -0.53*, -0.32) & (-0.45*, -0.45*, -0.12) & (-0.68*, -0.12, -0.51*) & (-0.60*, -0.26, -0.62*) & (-0.52*, -0.33, -0.21) & (0.01, -0.33, -0.26) \\
 & KLR\_Polynomial & (-0.38, -0.53*, -0.32) & - & (0.69*, 0.16, -0.2) & (0.4, 0.19, 0.25) & (0.44*, 0.29, -0.04) & (0.21, 0.26, 0.07) & (0.43*, 0.26, -0.26) \\
 & LSIF & (-0.45*, -0.45*, -0.12) & (0.69*, 0.16, -0.2) & - & (0.64*, 0.24, 0.11) & (0.51*, 0.17, -0.06) & (0.1, 0.37, -0.19) & (0.26, 0.37, -0.05) \\
 & uLSIF $\alpha$ = 0.25 & (-0.68*, -0.12, -0.51*) & (0.4, 0.19, 0.25) & (0.64*, 0.24, 0.11) & - & (0.74*, 0.27, 0.49*) & (0.51*, 0.72*, -0.28) & (-0.07, 0.72*, -0.33) \\
 & uLSIF $\alpha$ = 0.5 & (-0.60*, -0.26, -0.62*) & (0.44*, 0.29, -0.04) & (0.51*, 0.17, -0.06) & (0.74*, 0.27, 0.49*) & - & (0.42, 0.44*, 0.15) & (0.12, 0.44*, 0.22) \\
 & uLSIF $\alpha$ = 0.75 & (-0.52*, -0.33, -0.21) & (0.21, 0.26, 0.07) & (0.1, 0.37, -0.19) & (0.51*, 0.72*, -0.28) & (0.42, 0.44*, 0.15) & - & (0.03, 1.0*, 0.72*) \\
 & uLSIF $\alpha$ = 1 & (0.01, -0.33, -0.26) & (0.43*, 0.26, -0.26) & (0.26, 0.37, -0.05) & (-0.07, 0.72*, -0.33) & (0.12, 0.44*, 0.22) & (0.03, 1.0*, 0.72*) & - \\ \midrule
\end{tabular}%
}
\end{table*}

\begin{figure}[t!]
\centering
\includegraphics[width=0.99\columnwidth]{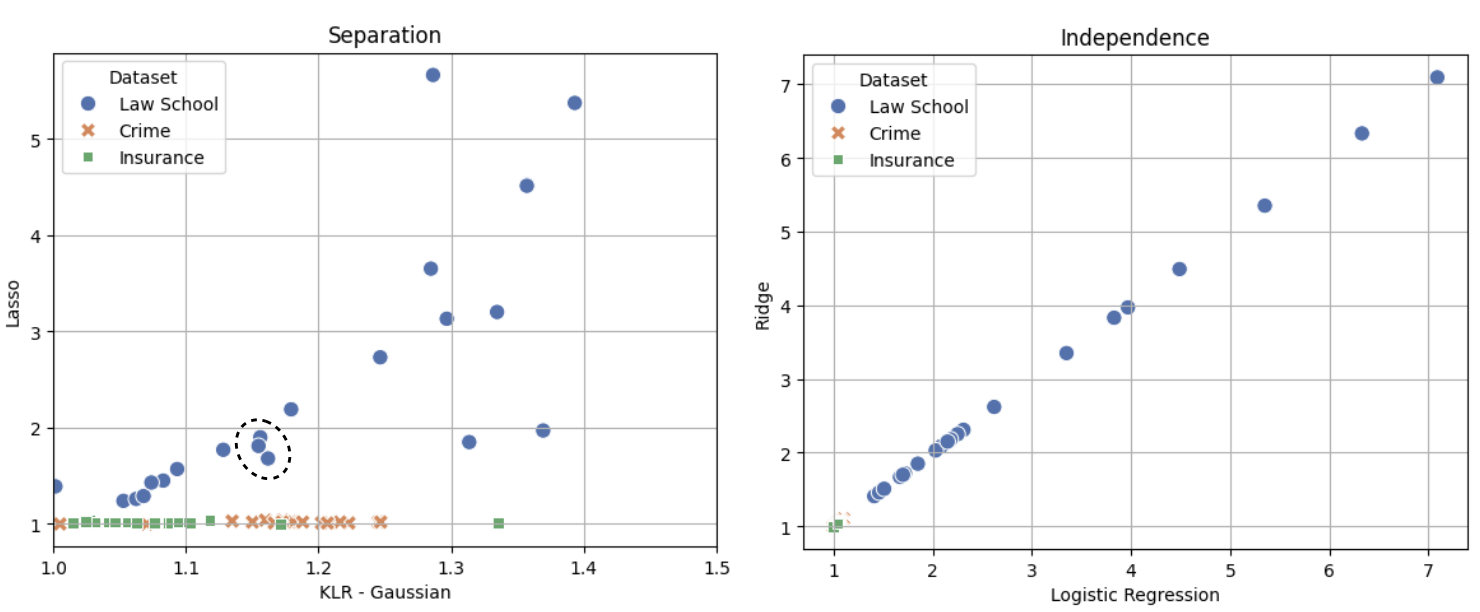} 
\caption{Scatter plots between Kernel Logistic Regression with Gaussian kernel and Lasso based methods approximating the Separation fairness metric (left sub-figure) and between Logistic Regression and Ridge based methods approximating the Independence metric (right sub-figure).}
\label{fig:fig1}
\end{figure}

To obtain further insights, we visualize the output of various fairness measurement methods. Figure \ref{fig:fig1} illustrates two cases: (1) The right plot shows the fairness values for Independence approximated by two methods: Logistic Regression and Ridge. This plot exhibits a positive and linear trend where an increase in fairness value in one model corresponds to an increase in the other model. However, this behavior is not consistent across all classifiers. (2) The left plot shows the estimated Separation values returned by methods using Kernel Logistic Regression (with Gaussian kernel) and Lasso.  We observe various anomalies where the fairness metric increases in one model but decreases in the other. This discrepancy is more obvious in the Law School dataset, where fairness metrics initially correlate but then begin to show inverse correlation in certain cases.

The high correlation of fairness values returned by methods with non-kernel-based classifier cores (i.e., Logistic Regression, Lasso, and Ridge) is reasonable because (1) the input of these classifiers is either a vectors of length one (corresponding to $Y_{pred}$ for measuring Independence) or a vector of length two (corresponding to $Y_{pred}$ and $Y_{true}$ for measuring Separation/Sufficiency), and (2) Lasso/Ridge only introduce a regularizer (L1/L2) to Logistic Regression, and given that the number of parameters to be learned are already small (because of small number of inputs), the use of regularization (which mainly aims to reduce the number of parameters) could result in intangible change in the resulted linear classifier.

Additionally, the low correlation between the fairness values returned by kernel-based classifiers and non-kernel ones could be justified as follows: kernel-trick technically helps converting those inherently linear classifiers to find non-linear classification boundaries, perhaps this could result in different output probability than the linear ones. Accordingly, these differences could change the measurement of fairness metrics relying on such approaches.


\section{Sensitivity of Fairness Measurements to the Choice of Density Ratio Estimation Approach}
The concerning results in the previous section have inspired us to expand the range of density ratio estimation methods for comprehensive sensitivity evaluation. In particular, we investigate the sensitivity of fairness measurement methods when other density ratio estimation approaches (other than the Probabilistic Classification-based ones) are used and how the choice of density ratio estimation method affects the resulting fairness values. Next sub-sections describe details of our methodology and obtained results.
\subsection{Methodology}

We follow the same technical steps as mentioned in the previous section. However, we implement several ratio matching approaches (rather than the probabilistic classifiers), including Least-Squares Importance Fitting (LSIF) and its variant, i.e., unconstrained Least-Squares Importance Fitting (uLSIF) \cite{18-kanamori2009least}, to examine the impact of replacing the proposed Probabilistic Classification approach with these methods on measuring Independence, Separation, and Sufficiency fairness metrics. These methods estimate the density ratio directly by framing it as a least-squares function fitting problem; especially, the uLSIF variant offers improvements such as enhanced numerical stability and computational efficiency, particularly in model selection scenarios \cite{18-kanamori2009least}. Our code base is available at \url{https://github.com/wahab1412/Fairness_measurement}.

\subsection{Experimental Results}
Table \ref{dens_results_sep} shows the results of our evaluations for Independence, Separation, and Sufficiency metrics.
The results show a generally low correlation when comparing the ratio matching-based approach with the Probabilistic Classification-based ones. For example, the Spearman correlation between the outcome of methods relying on Logistic Regression and uLSIF $\alpha$=1 are (-0.30, 0.38, 0.02). These confirm the sensitivity of density ratio-based fairness measurement methods to the choice of density ratio estimation approach. 

Additionally, the output of ratio matching approaches seem to show a low level of correlation in some cases. For example, the correlation between the output of methods relying on LSIF and uLSIF $\alpha$=1 are (-0.28, 0.37, 0.08) and (0.26, 0.37, -0.05) when approximating Independence and Sufficiency, respectively. Notably, there is significant variation in correlation values across datasets, even within the same methods. For example, when measuring Sufficiency, Spearman correlations between the fairness values returned by methods with uLSIF $\alpha$ = 1 and uLSIF $\alpha$ = 0.25 cores are (-0.07, 0.72*, -0.33), which show a large variability  across the three datasets. Similarly, the correlation between the Sufficiency values estimated by methods that have uLSIF $\alpha$ = 1 and uLSIF $\alpha$ = 0.75 as their core algorithm is (0.03, 1.00*, 0.72*), which show significant variation as well.

\begin{figure}[h!]
\centering
\includegraphics[width=0.4\textwidth]{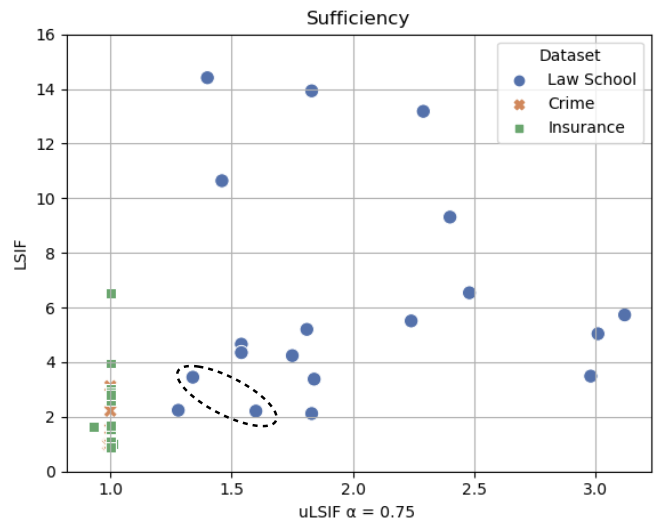} 
\caption{A scatter plot between two output of two fairness measurement methods relying on LSIF and uLISF $\alpha$ = 0.75 (when approximating the Sufficiency fairness metric across three datasets).}
\label{fig:fig2}
\end{figure}

Visual representation of the results further supports these observations. Figure \ref{fig:fig2} compares LSIF and uLISF $\alpha$ = 0.75 for measuring Sufficiency metric and demonstrates no clear trend or correlation, reinforcing the inconsistencies in fairness measurements across different estimation methods, which is in line with the findings of the previous section and \cite{15-almajed2024consistency}.

In summary, this compressive analysis shows that approximating fairness metrics through density ratio estimation methods is highly sensitive, depending heavily on the chosen approach and classifier within the probabilistic classification framework. As a result, these methods with various approximation cores may yield contradictory insights regarding the relative fairness of machine learning models, similar to \cite{15-almajed2024consistency}. Such inconsistencies could have a profound effect on fair regression research and its practical applications as fairness measurement in a primary step in ensuring fairness. For example, ML models may produce inaccurate low-risk assessments for patients from marginalized communities or predict higher probabilities of loan default for certain demographic groups. Therefore, similar to recent research \cite{15-almajed2024consistency}, we also emphasize the need for further research to develop better methods for assessing fairness metrics in the regression problem domains and address the discovered sensitivity and inconsistency issues.

\begin{table*}[]
\caption{Spearman correlations between pairs of methods approximating the Independence fairness metric at different mean intervals.}
\label{tab:corr_interval_0_updated}
\centering
\resizebox{0.87\textwidth}{!}{%
\begin{tabular}{lllllll}
\hline
Mean Interval & Fairness Measurement & Logistic Regression & Ridge & \multicolumn{1}{r}{Lasso} & KLR\_Gaussian & KLR\_Polynomial \\ \hline
0 - 0.9 & Logistic Regression & - & 1.00 & 1.00 & 0.85 & 0.94 \\
 & Ridge & 1.00 & - & 0.99 & 0.85 & 0.94 \\
 & Lasso & 1.00 & 0.99 & - & 0.84 & 0.93 \\
 & KLR\_Gaussian & 0.85 & 0.85 & 0.84 & - & 0.70 \\
 & KLR\_Polynomial & 0.94 & 0.94 & 0.93 & 0.70 & - \\ \hline
1 - 1.9 & Logistic Regression & - & 1.00 & 1.00 & 0.96 & 0.88 \\
 & Ridge & 1.00 & - & 1.00 & 0.96 & 0.88 \\
 & Lasso & 1.00 & 1.00 & - & 0.96 & 0.88 \\
 & KLR\_Gaussian & 0.96 & 0.96 & 0.96 & - & 0.92 \\
 & KLR\_Polynomial & 0.88 & 0.88 & 0.88 & 0.92 & - \\ \hline
2 - 2.9 & Logistic Regression & - & 0.99 & 1.00 & 0.92 & 0.64 \\
 & Ridge & 0.99 & - & 0.99 & 0.87 & 0.62 \\
 & Lasso & 1.00 & 0.99 & - & 0.92 & 0.64 \\
 & KLR\_Gaussian & 0.92 & 0.87 & 0.92 & - & 0.55 \\
 & KLR\_Polynomial & 0.64 & 0.62 & 0.64 & 0.55 & - \\ \hline
3 - 3.9 & Logistic Regression & - & 0.89 & 0.99 & 0.82 & -0.12 \\
 & Ridge & 0.89 & - & 0.94 & 0.54 & 0.21 \\
 & Lasso & 0.99 & 0.94 & - & 0.77 & -0.05 \\
 & KLR\_Gaussian & 0.82 & 0.54 & 0.77 & - & -0.35 \\
 & KLR\_Polynomial & -0.12 & 0.21 & -0.05 & -0.35 & - \\ \hline
\end{tabular}%
}
\end{table*}

\section{Potential Reason of Inconsistencies}
The results of our experiments, in particular the Probabilistic Classification-based approaches, show that relative fairness of various ML models can change significantly when we replace the core probabilistic classifier, and these inconsistencies can differ across datasets. One potential factor that may attribute to these observations could be the sensitivity of these approaches to the high probability values predicted by the underlying classifier. For example, as mentioned before, Independence is defined as $\frac{P(A = 1 \mid Y_{pred}) \cdot P(A = 0)}{P(A = 0 \mid Y_{pred}) \cdot P(A = 1)}$ \cite{7-steinberg2020fairness}, in which the nominator ($P(A = 1 \mid Y_{pred})$) is approximated using a probabilistic classifier (such as Logistic Regression), and the denominator value is derived from the nominator value as follows: $P(A = 0 \mid Y_{pred})=1-P(A = 1 \mid Y_{pred})$. Then, assuming the same size for privileged and unprivileged groups, if the classifier's predicted probability for  $P(A = 1 \mid Y_{pred})$ changes from 0.9990 to 0.9999 for a particular data point, that ratio will increase from 999 to 9999 (which is an approximately 10 times increase). Such sudden changes of ratios for just a few data points could significantly affect the overall Independence value (computed through averaging such ratios across all data points in a dataset), and accordingly, the consistency of fairness values returned by various approximation methods.

\begin{figure}[h!]
\centering
\includegraphics[width=0.4\textwidth]{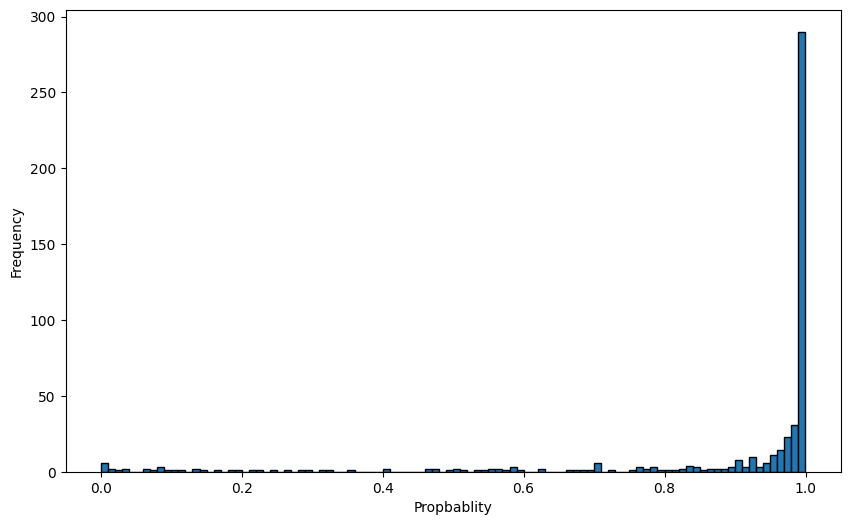} 
\caption{A histogram of probabilities predicted by Logistic Regression for computing the Independence fairness metric for the Ridge regression algorithm on the Communities and Crime dataset.}
\label{fig:fighistorigram}
\end{figure}

To be more specific, we observe high predicted probabilities for $P(A = 1 \mid Y_{pred})$ in our real-world datasets. Figure \ref{fig:fighistorigram} shows the distribution of Logistic Regression's predicted probabilities when computing Independence fairness 
on the Communities and Crime dataset. This figure shows that, for many data points, the predicted probability is high; for example, for 112 datapoints, the predicted probability is above 0.999. In this situation, the estimated Independence value is 79.78. However, when we impose a threshold on the number of decimal points for the predicted probability (as a post-processing step), the Independence values change significantly; for example, the Independence values will change to 46.17, 8.19, and 0.94 when setting the maximum threshold to 0.999, 0.99, and 0.9, respectively. This could show the extent to which these fairness measurement methods are sensitive to a minor variations in the high probability values predicted by the underlying probabilistic classifiers.

One potential factor behind such high probabilities is the distribution of privileged and unprivileged groups and their level of overlaps. In particular, when there is a large gap between samples of these two classes, it is highly likely that the underlying probabilistic classifier predicts a high probability for certain samples belonging to the privileged group, which could significantly affect the fairness values accordingly. In those cases, most likely, high inconsistencies would be observed among the fairness values computed through using various core probabilistic classifiers. We validate this hypothesis by generating synthetic datasets in which levels of overlap among samples of privileged and unprivileged classes varies across datasets.

\begin{figure}[]
\centering
\includegraphics[width=0.48\textwidth]{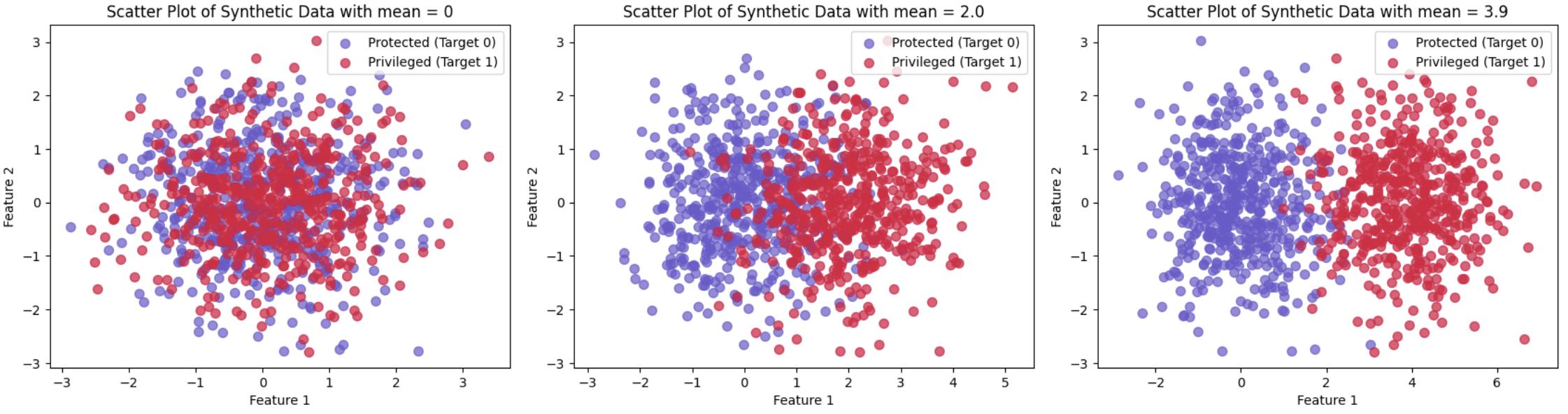} 
\caption{Scatter plots of the synthetic data at three mean values (0, 2, and 3.9), illustrating how varying overlap influences class separability.}
\label{fig:fig_2d}
\end{figure}

To evaluate the impact of data distribution on fairness values, we created 40 synthetic datasets each containing 1,000 data points evenly split between the privileged and unprivileged classes. To control the data distribution, samples for the unprivileged group are generated from a normal distribution with a mean of 0 and a standard deviation of 1. These samples are then combined with samples generated for the privileged class to form a single dataset with one feature (Feature 1).
For the privileged class, we iteratively generated samples with the same standard deviation of 1 but an incrementally increasing mean. Starting with a mean of 0, we increased the mean by 0.1 for each subsequent dataset. This approach allowed us to gradually vary the degree of overlap between the two classes across datasets.
For example, in the first dataset, both classes are sampled from the same distribution (normal distribution with mean of 0 and standard deviation of 1), resulting in equal probabilities for each data point belonging to either class. However, as the mean of the privileged class increased while keeping the mean of the unprivileged class constant to 0, the overlap between the two groups is reduced, creating datasets with distinct probabilities for each point belonging to either group.
Moreover, examining how data distribution affects Separation fairness metrics requires two-dimensional data (predicted versus ground-truth values). In our synthetic data generation process, we have already created Feature 1 with varying degrees of overlap. For Feature 2, both classes were sampled from a normal distribution with a mean of 0 and a standard deviation of 1. Figure \ref{fig:fig_2d} provides scatter plots illustrating the data distributions at different mean values. In the left plot (mean = 0), the classes overlap entirely. As the mean increases to 2 (middle plot), overlap decreases, making the classes more separable. At a mean of 3.9 (right plot), there is minimal overlap between the two classes.
Through this process, we created 40 datasets with varying degrees of class overlap to analyze how classifiers respond to different distributions and levels of class overlap.

After generating the synthetic datasets, we proceed to our experiments, which consist of two main steps. In the first step, we compute the Spearman correlations between fairness values generated by various fairness measurement methods relying on different classifiers to investigate the consistency of various fairness measurement methods under different data distribution. Then, the second step involves imposing a threshold on high predicted probabilities produced by these classifiers, and then, comparing the correlation values before and after thresholding to determine its impact on the consistency of fairness measurement methods. The outcomes of these two experimental steps are detailed in the following sub-sections.

\subsection{Step 1. Evaluation of the Consistency of Fairness Measurement Methods on Synthetic Data}
In this step, the fairness metrics were calculated for each of the 40 datasets using different fairness measurement methods with various underlying core classifiers. This resulted in 40 fairness values per each fairness measurement method. To analyze the results, we aggregated the fairness values computed for 10 synthetic datasets within specific mean intervals of the privileged class distribution. This grouping resulted in four mean intervals: [0--0.9], [1--1.9], [2--2.9], and [3--3.9], covering the total 40 datasets. Then, for each interval, we calculated the Spearman correlation of the fairness values for each pair of methods.

Table \ref{tab:corr_interval_0_updated} summarizes the results of this experiment, where we computed the correlation values for each pair of fairness measurements (relying on different classifiers) across different ranges of mean intervals.
The overall trend suggests that as the mean value increases (reducing overlap between the two classes), the correlation between fairness values produced by different classifiers decreases.
For example, within the mean interval of [0--0.9], the fairness values exhibit consistently high correlations across all classifiers. In particular, Logistic Regression, Ridge, and Lasso showed a perfect correlation of 1. KLR\_Polynomial demonstrated a high correlation of 0.94 with Logistic Regression, Ridge, and Lasso, and KLR\_Gaussian exhibited a correlation of 0.85 with these classifiers, except for KLR\_Polynomial, where the correlation value was 0.7.
However, as the mean interval increases, the correlations tends to show an opposite trend. For instance, the correlation between KLR\_Polynomial and Logistic Regression dropped from 0.94 to -0.12, while the correlation between KLR\_Polynomial and KLR\_Gaussian decreased from 0.70 to -0.35 between the first and last mean intervals. Similarly, KLR\_Gaussian exhibited a correlation of 0.54 with Ridge at the last interval. These findings suggest that as the overlap between the two classes (privileged and unprivileged) decreases, the consistency of fairness measurements across different classifiers could diminish.

\begin{figure*}[]
\centering
\includegraphics[width=1\textwidth]{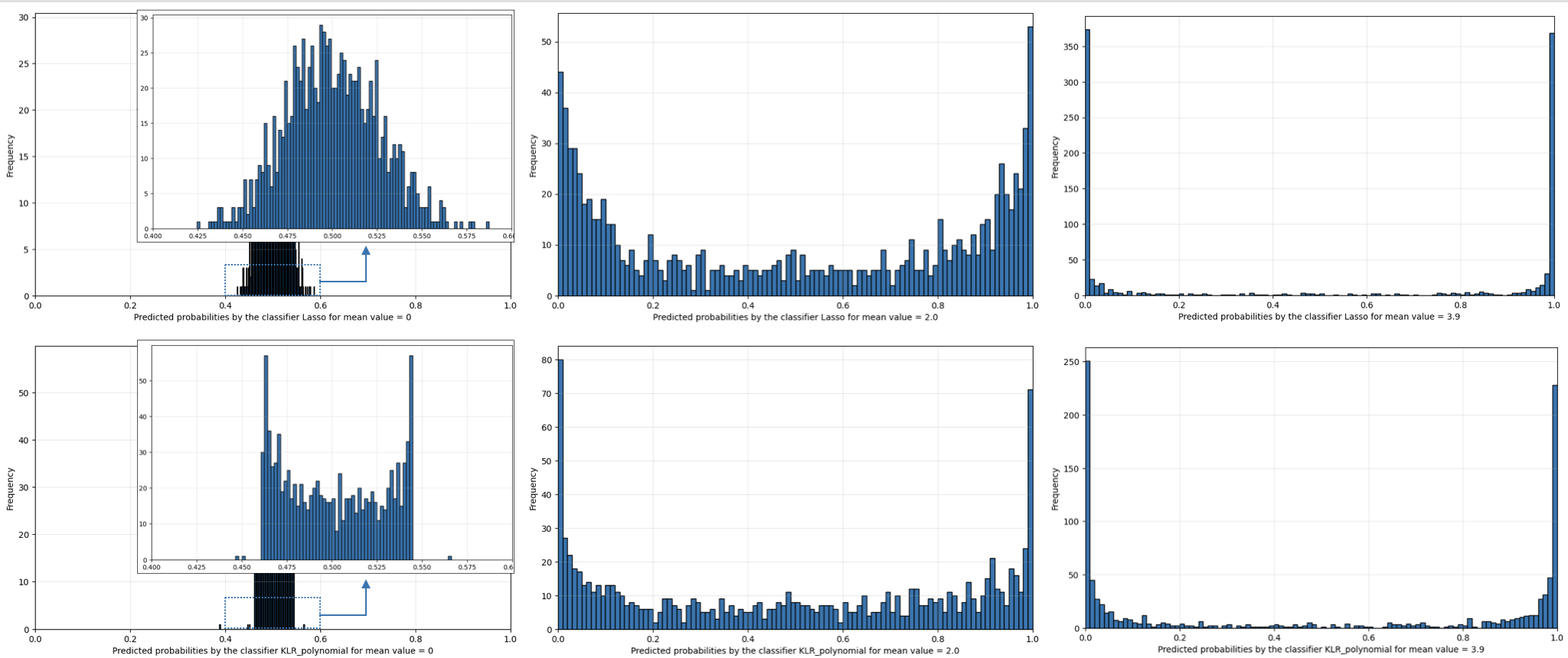} 
\caption{Histograms of the predicted probabilities from Lasso (top row) and Kernel Logistic Regression with a Polynomial kernel (bottom row), evaluated on synthetic datasets with different mean values (0, 2, and 3.9).}
\label{fig:fig_props}
\end{figure*}

Moreover, when we analyze the output of these classifiers more closely, we observe a trend similar to our findings in the real dataset. In some cases, classifiers assign very high probabilities to many data points. Generally, when the mean value for the sampled data in the synthetic datasets is low (indicating significant overlap between classes), most classifiers tend to produce probabilities concentrated around 0.5. This outcome is expected because both the privileged and unprivileged groups are drawn from closely overlapping distributions, making it difficult for the models to distinguish between them. As the mean for the privileged group distribution increases, the overlap diminishes, and a broader range of probability values emerges.

Figure \ref{fig:fig_props} illustrates the predicted probability distributions for Lasso (top row) and Kernel Logistic Regression with a Polynomial kernel (bottom row) across three different mean values: 0, 2, and 3.9. In the leftmost plots, where the mean is 0, the probabilities cluster around 0.5 because of the high degree of overlap between the two groups, which hinders classification. In contrast, the middle plots (mean = 2) show that around 50 data points are predicted with probabilities exceeding 0.9, indicating increased classifier probability as overlap decreases. Finally, when the mean value is raised to 3.9 (rightmost plots), the distributions concentrate near the edges, indicating that many data points receive very high predicted probabilities due to minimal overlap between the two privileged and unprivileged groups.

Furthermore, using this synthetic data generation approach we can calculate the actual fairness value for Independence because exact probability density functions (PDF) for the nominator and denominator is known. 
This allows us to compare the actual fairness of the dataset with the fairness results produced by various classifiers, providing a benchmark for evaluating fairness estimation methods' performance. Table \ref{tab:actuall_updated} shows the Spearman correlations between the actual fairness values and a Probabilistic Classification-based one with different core classifiers across different ranges of mean intervals. According to the results, inconsistencies increase as the mean value rises. For example, when the mean interval range is [0--0.9], the average correlation is 0.94; however, this correlation decreases to 0.66 when the mean interval range reaches [3--3.9].

\begin{table}[]
\caption{Spearman correlations between actual and estimated Independence metric (with various core probabilistic classifiers) across various mean intervals.}
\label{tab:actuall_updated}
\centering
\resizebox{0.48\textwidth}{!}{%
\begin{tabular}{llrrrrr}
\hline
 Mean Interval & \multicolumn{1}{l}{Logistic Regression} & \multicolumn{1}{l}{Ridge} & \multicolumn{1}{l}{Lasso} & \multicolumn{1}{l}{KLR\_Gaussian} & \multicolumn{1}{l}{KLR\_Polynomial} \\ \hline
 0-0.9& 0.98 & 0.98 & 0.98 & 0.81	 & 0.95 \\
  1-1.9 & 1.00 & 1.00 & 1.00 & 0.96 & 0.88 \\
  2-3.9 & 0.98 & 0.99 & 0.98 & 0.88 & 0.65 \\
   3-3.9 & 0.96 & 0.78 & 0.93 & 0.85 & -0.21 \\
\hline

\end{tabular}%
}
\end{table}

\begin{table*}[h!]
\caption{Spearman correlations between pairs of methods approximating the Separation fairness metric at different mean intervals.}
\label{tab:2d_sep_Spearman_updated}
\centering
\resizebox{0.9\textwidth}{!}{%
\begin{tabular}{lllllll}
\hline
Mean Interval & Fairness Measurement & Logistic Regression & Ridge & \multicolumn{1}{r}{Lasso} & KLR\_Gaussian & KLR\_Polynomial \\ \hline
0 - 0.9 & Logistic Regression & - & 1.00 & 1.00 & 0.84 & 0.92 \\
 & Ridge & 1.00 & - & 1.00 & 0.84 & 0.93 \\
 & Lasso & 1.00 & 1.00 & - & 0.84 & 0.93 \\
 & KLR\_Gaussian & 0.84 & 0.84 & 0.84 & - & 0.84 \\
 & KLR\_Polynomial & 0.92 & 0.93 & 0.93 & 0.84 & - \\ \hline
1 - 1.9 & Logistic Regression & - & 1.00 & 1.00 & 0.99 & 0.89 \\
 & Ridge & 1.00 & - & 1.00 & 0.99 & 0.89 \\
 & Lasso & 1.00 & 1.00 & - & 0.99 & 0.89 \\
 & KLR\_Gaussian & 0.99 & 0.99 & 0.99 & - & 0.88 \\
 & KLR\_Polynomial & 0.89 & 0.89 & 0.89 & 0.88 & - \\ \hline
2 - 2.9 & Logistic Regression & - & 0.99 & 1.00 & 0.95 & 0.48 \\
 & Ridge & 0.99 & - & 0.99 & 0.94 & 0.56 \\
 & Lasso & 1.00 & 0.99 & - & 0.95 & 0.48 \\
 & KLR\_Gaussian & 0.95 & 0.94 & 0.95 & - & 0.54 \\
 & KLR\_Polynomial & 0.48 & 0.56 & 0.48 & 0.54 & - \\ \hline
3 - 3.9 & Logistic Regression & - & 0.89 & 0.99 & 0.76 & 0.09 \\
 & Ridge & 0.89 & - & 0.94 & 0.49 & 0.47 \\
 & Lasso & 0.99 & 0.94 & - & 0.72 & 0.18 \\
 & KLR\_Gaussian & 0.76 & 0.49 & 0.72 & - & -0.35 \\
 & KLR\_Polynomial & 0.09 & 0.47 & 0.18 & -0.35 & - \\ \hline
\end{tabular}%
}
\end{table*}

Overall, the correlation values tend to be high when class overlap is significant.
Table \ref{tab:2d_sep_Spearman_updated} displays the Spearman correlation values among classifiers approximating the Separation metric across different mean intervals. For example, during the lower mean intervals ([0--0.99] and [1--1.9]), correlation values among all classifiers exceed 0.84, indicating strong agreement. However, as the mean interval increases, the correlations decreases significantly. In the interval of [3--3.9], the Spearman correlation between Kernel Logistic Regression with a Polynomial kernel and other classifiers--such as Logistic Regression, Ridge, Lasso, and Kernel Logistic Regression with a Gaussian kernel--drops to 0.09, 0.47, 0.18, and --0.35, respectively. 
These results highlight how varying degrees of class overlap in data distributions can substantially influence correlation values when evaluating the Separation fairness metric.

\begin{table*}[]
\caption{Spearman correlations for the Independence fairness metric, computed using various measurement methods (relying on different classifiers), before (top) and after (bottom) thresholding their predicted probabilities.}
\label{tab:spear_threshold_updated}
\resizebox{\textwidth}{!}{%
\begin{tabular}{lllllll}
\hline
Corr. Metric &  & Logistic Regression & Ridge & Lasso & KLR\_Gaussian & KLR\_Polynomial \\ \hline
Spearman & Logistic Regression & - & 0.89 & 0.99 & 0.82 & -0.12 \\
(before thresholding) & Ridge & 0.89 & - & 0.94 & 0.54 & 0.21 \\
 & Lasso & 0.99 & 0.94 & - & 0.77 & -0.05 \\
 & KLR\_Gaussian & 0.82 & 0.54 & 0.77 & - & -0.35 \\
 & KLR\_Polynomial & -0.12 & 0.21 & -0.05 & -0.35 & - \\ \hline
Spearman & Logistic Regression & - & 1.00 & 1.00 & 0.90 & 0.98 \\
(after thresholding) & Ridge & 1.00 & - & 1.00 & 0.90 & 0.98 \\
 & Lasso & 1.00 & 1.00 & - & 0.90 & 0.98 \\
 & KLR\_Gaussian & 0.90 & 0.90 & 0.90 & - & 0.95 \\
 & KLR\_Polynomial & 0.98 & 0.98 & 0.98 & 0.95 & - \\ \hline
 
\end{tabular}%
}
\end{table*}

\subsection{Step 2. Investigating the Impact of Thresholding}
Because this fairness measurement method tend to be highly sensitive to extreme predictions, the distribution of predicted probabilities plays a critical role in determining the overall fairness score. We noticed that some classifiers produce exceptionally high probabilities, which can disproportionately skew fairness values and, in turn, influence the correlation among the fairness measurement methods. To address this issue, we investigate the effect of imposing a threshold on predicted probabilities by comparing the correlation values among methods relying on different classifiers before and after applying the threshold.

We apply a threshold of 0.99 to all probabilities predicted by each classifier, then recompute the Independence fairness metric. Next, we calculate the Spearman correlation values among the classifiers to observe how thresholding influences their correlations, particularly within the mean interval of [3--3.9], which exhibited the lowest correlation values prior to thresholding. 
Table \ref{tab:spear_threshold_updated} presents the Spearman correlation values before and after thresholding, revealing that correlations generally improve across all pairs of methods, especially those involving Kernel Logistic Regression (KLR). For instance, the correlation between Lasso and KLR with Gaussian kernel increases from 0.54 to 0.90, and from 0.21 to 0.98 with the Polynomial kernel. Likewise, the correlation between the two KLR classifiers rises from -0.35 to 0.95. 
These findings suggest that thresholding can help reduce inconsistencies in synthetic datasets when measuring the Independence fairness metric.

\begin{table*}[]
\caption{Spearman correlations for the Separation fairness metric, computed using various measurement methods (relying on different classifiers), before (top) and after (bottom) thresholding their predicted probabilities.}
\label{tab:Spearman_sep_thresh_updated}
\resizebox{\textwidth}{!}{%
\begin{tabular}{lllllll}
\hline
Corr. Metric &  & Logistic Regression & Ridge & Lasso & KLR\_Gaussian & KLR\_Polynomial \\ \hline
Spearman & Logistic Regression & - & 0.89 & 0.99 & 0.76 & 0.09 \\
(before thresholding) & Ridge & 0.89 & - & 0.94 & 0.49 & 0.47 \\
 & Lasso & 0.99 & 0.94 & - & 0.72 & 0.18 \\
 & KLR\_Gaussian & 0.76 & 0.49 & 0.72 & - & -0.35 \\
 & KLR\_Polynomial & 0.09 & 0.47 & 0.18 & -0.35 & - \\ \hline
Spearman & Logistic Regression & - & 1.00 & 1.00 & 0.88 & 0.71 \\
(after thresholding) & Ridge & 1.00 & - & 1.00 & 0.88 & 0.71 \\
 & Lasso & 1.00 & 1.00 & - & 0.88 & 0.71 \\
 & KLR\_Gaussian & 0.88 & 0.88 & 0.88 & - & 0.47 \\
 & KLR\_Polynomial & 0.71 & 0.71 & 0.71 & 0.47 & - \\ \hline
\end{tabular}%
}
\end{table*}

We also investigate the impact of applying a threshold of 0.99 to predicted probabilities used in calculating the Separation fairness metric. The results suggest that thresholding generally enhances correlation values between these types of fairness measurement methods. Similarly, we focus on the mean interval of [3--3.9], as it showed the lowest correlation values before thresholding was applied. Table \ref{tab:Spearman_sep_thresh_updated} compares Spearman correlation values before and after thresholding. As the results show, notable improvements are observed even when the initial correlation values are very low. For example, the correlation between Kernel Logistic Regression with Gaussian and Polynomial kernels increases significantly from -0.35 to 0.47 after thresholding. High correlations also experience improvement; the correlation between Logistic Regression and Ridge classifiers rises from 0.89 to 1.00. 
The findings from the experiments conducted on synthetic datasets underline that applying a threshold can effectively improve correlation values in evaluating the Separation fairness metric, thereby mitigating the influence of extreme predicted probabilities.

After observing that imposing a threshold on synthetic data improves overall consistency, we now examine its impact on our real-world datasets--specifically, the Law School, Communities and Crime, and Insurance datasets. We apply a threshold of 0.99 to predicted probabilities and calculate correlation values before and after thresholding. The results present a mixed picture, with thresholding improving correlations in some cases while leaving them unchanged or even reducing them in others.
For instance, Spearman correlation values between Logistic Regression with Gaussian and Polynomial kernels, measuring the Independence fairness metric, initially stand at (0.1, -0.58, -0.67) for Law School, Communities and Crime, and Insurance datasets, respectively. After thresholding, these correlations change to (-0.21, -0.58, -0.67), showing no change for the Communities and Crime and Insurance datasets, but indicating a negative effect for the Law School dataset, where the correlation drops from 0.1 to -0.21.
In contrast, thresholding can also lead to improvements. For example, the correlation between Lasso and Kernel Logistic Regression with a Gaussian kernel improves from 0.53 to 0.64 for the Communities and Crime dataset, while remaining unchanged for the other datasets.
These results from experiments on real-world datasets suggest that imposing thresholds to predicted probabilities can \textit{occasionally} (not always) reduce observed inconsistencies, which in turn, suggests that factors beyond high probability values may contribute to inconsistencies, requiring further investigation.

In summary, the results of our analysis on the real-world and synthetic data show that the fairness metrics computed through Probabilistic Classification-based approach could be highly sensitive to the performance of the underlying classifier, which in turn, could be affected by the data distribution. Particularly, a major gap between some data points of privileged and unprivileged groups could affect the consistency of the outcome of various fairness estimation approaches with different underlying probabilistic classifiers.

\section{Conclusion}
This paper seeks to address the gap in investigating the sensitivity of density-ratio estimation approaches for approximating fairness notions through analyzing the impact of changing the underlying density-ratio estimation approaches (such as various Probabilistic Classification-based ones and ratio matching approaches) on the output fairness measures. The findings reveal that the choice of underlying density-ratio estimation could significantly impact the fairness measurement results, and even, result in contradictory results with respect to the relative fairness of ML models. This variation is clearly demonstrated in our analysis of LSIF and uLSIF ($\alpha$ = 0.5) approaches, which showed different correlation values of (0.46, -0.02, -0.02) across three regression tasks. 
Furthermore, this paper provides insights on the potential reasons of some inconsistencies through generating and analyzing synthetic data with various distributions. The empirical results suggest that the distribution of privileged and unprivileged samples and their overlap could affect the fairness estimation process and consistency of the outcome of various fairness measurement methods in regression.

\bibliography{aaai25}
\newpage
\appendix
\section{Appendix}
For the purpose of completeness, in this section, we present the Kendall correlation values for all experimental results. Overall, we observe a similar pattern between the Spearman and Kendall correlation analyses, with both showing either strong or weak correlations in the same instances.

Table \ref{tab:indep_q1_updated_k} reports the Kendall correlations for Independence, Separation, and Sufficiency across the three datasets for Probabilistic Classification-based fairness estimation methods with various probabilistic classifier cores. Overall, the findings are similar to those suggested by Spearman; i.e., some methods such as Logistic Regression, Ridge, and Lasso, show consistently high correlations, whereas kernel-based approaches often display low consistencies. For example, the correlations between Ridge and KLR-Polynomial for the Independence metric are (0.25, -0.08, -0.01) across the three datasets.

\begin{table*}[h]
\caption{Kendall correlations between the output of pairs of fairness measurement methods with various underlying probabilistic classifiers when measuring Independence, Separation and Sufficiency (shown in each column/row). Results are presented across Law School, Communities and Crime, and Insurance datasets, respectively.}
\label{tab:indep_q1_updated_k}
\resizebox{\textwidth}{!}{%
\begin{tabular}{lllllll}
\hline
Corr. Metric &  & Logistic Regression & Ridge & Lasso & KLR\_Gaussian & KLR\_Polynomial \\ \hline

Independence & Logistic Regression & - & (1.00*, 1.00*, 1.00*) & (1.00*, 0.90*, 0.91*) & (0.73*, 0.06, -0.39*) & (0.11, -0.29, 0.07) \\
 & Ridge & (1.00*, 1.00*, 1.00*) & - & (1.00*, 0.93*, 0.89*) & (0.76*, 0.07, -0.31) & (0.25, -0.08, -0.01) \\
 & Lasso & (1.00*, 0.90*, 0.91*) & (1.00*, 0.93*, 0.89*) & - & (0.77*, 0.02, -0.29) & (0.26, -0.1, -0.01) \\
 & KLR\_Gaussian & (0.73*, 0.06, -0.39*) & (0.76*, 0.07, -0.31) & (0.77*, 0.02, -0.29) & - & (0.21, 0.45*, -0.47*) \\
 & KLR\_Polynomial & (0.11, -0.29, 0.07) & (0.25, -0.08, -0.01) & (0.26, -0.1, -0.01) & (0.21, 0.45*, -0.47* & - \\ \hline
 Separation & Logistic Regression & - & (1.00*, 1.00*, 1.00*) & (1.00*, 0.93*, 0.89*) & (0.76*, 0.07, -0.31) & (0.25, -0.08, -0.01) \\
 & Ridge & (1.00*, 1.00*, 1.00*) & - & (1.00*, 0.93*, 0.89*) & (0.76*, 0.07, -0.31) & (0.25, -0.08, -0.01) \\
 & Lasso & (1.00*, 0.93*, 0.89*) & (1.00*, 0.93*, 0.89*) & - & (0.77*, 0.02, -0.29) & (0.26, -0.10, -0.01) \\
 & KLR\_Gaussian & (0.76*, 0.07, -0.31) & (0.76*, 0.07, -0.31) & (0.77*, 0.02, -0.29) & - & (0.21, 0.45*, -0.47*) \\
 & KLR\_Polynomial & (0.25, -0.08, -0.01) & (0.25, -0.08, -0.01) & (0.26, -0.10, -0.01) & (0.21, 0.45*, -0.47*) & - \\ \hline
 Sufficiency & Logistic Regression & - & (1.00*, 1.00*, 1.00*) & (0.98*, 1.00*, 0.87*) & (0.48*, 0.49*, 0.08) & (-0.28, -0.14, -0.32) \\
 & Ridge & (1.00*, 1.00*, 1.00*) & - & (0.98*, 1.00*, 0.87*) & (0.48*, 0.49*, 0.08) & (-0.28, -0.14, -0.32) \\
 & Lasso & (0.98*, 1.00*, 0.87*) & (0.98*, 1.00*, 0.87*) & - & (0.47*, 0.49*, 0.02) & (-0.28, -0.14, -0.28) \\
 & KLR\_Gaussian & (0.48*, 0.49*, 0.08) & (0.48*, 0.49*, 0.08) & (0.47*, 0.49*, 0.02) & - & (-0.59*, -0.32*, -0.64*) \\
 & KLR\_Polynomial & (-0.28, -0.14, -0.32) & (-0.28, -0.14, -0.32) & (-0.28, -0.14, -0.28) & (-0.59*, -0.32*, -0.64*) & - \\ \hline
\end{tabular}%
}
\end{table*}

The Kendall results for studying the sensitivity of fairness measurement methods when utilizing density ratio estimation approaches beside the proposed Probabilistic Classification-based ones are presented in Table \ref{dens_results_ind_k}. Overall, we observe a high degree of inconsistency across different measurement approaches. For instance, the correlation between Logistic Regression and LSIF for measuring the Sufficiency metric are (-0.32*, -0.38*, -0.04) across the three datasets. Moreover, the level of inconsistency between the same pair of measurement methods varies across datasets. For example, the correlation between uLSIF $\alpha$ = 0.25 and LSIF when measuring the Independence metric is (0.65*, 0.24, 0.06).

We also conduct a Kendall correlation analysis on the generated synthetic datasets, aggregating results for every 10 datasets within specific mean intervals, resulting in four distinct intervals. Table \ref{tab:Kendall_intervall_updated_k} shows the Kendall correlation between different measurement methods approximating the Independence fairness metric at different mean intervals.  The Kendall analysis shows a similar pattern to the Spearman analysis, where an increase in the mean value corresponded to a decrease in correlation. For example, the correlation between Lasso and KLR\_Polynomial drops from 0.82 to 0.02 between the first and last mean intervals. A similar trend is observed when analyzing the Separation fairness metric. Table \ref{tab:2d_sep_Kendall_updated_k} presents the Kendall correlation values for Separation metric. It shows that correlations tend to decrease as the mean value increases. For instance, the correlation between KLR\_Polynomial and KLR\_Gaussian drops from 0.73 to -0.24 between the first and last mean intervals.

Moreover, we computed the Kendall correlation analysis between the predicted fairness metric and the actual fairness values of the synthetic datasets. Table \ref{tab:actuall_updated_k} presents the Kendall correlations between the predicted Independence values produced by Probabilistic Classification-based approaches and the actual Independence values across different mean intervals. Overall, inconsistencies appear to increase with higher mean values. For example, the average correlation decreases from 0.87 to 0.57 between the first and last mean intervals.

Furthermore, we study the impact of applying a threshold on the consistency of fairness measurement methods. Tables \ref{tab:Kendall_threshold_updated_k} and \ref{tab:Kendall_sep_thresh_updated_k} show the Kendall correlation values for the final mean interval before and after thresholding, for the Independence and Separation fairness metrics, respectively. The overall trend indicates that thresholding helps improve consistency across different fairness measurement methods on the synthetic datasets. For instance, the correlation between Ridge and KLR\_Gaussian increased from 0.33 to 0.78 for the Independence metric, and from 0.29 to 0.73 for the Separation metric.

In conclusion, the Kendall correlation analysis highlights the sensitivity of density ratio estimation approaches both in terms of the underlying classifiers used in Probabilistic Classification--based methods and the specific density ratio techniques employed. These findings align with and reinforce the conclusions drawn from the Spearman correlation analysis.

\begin{table*}[t]
\centering
\caption{Kendall correlations between the output of pairs of fairness measurement methods with various density ratio estimation approaches when measuring Independence, Separation and Sufficiency (shown in each column/row). Results are presented across Law School, Communities and Crime, and Insurance datasets, respectively.}
\label{dens_results_ind_k}
\resizebox{\textwidth}{!}{%
\begin{tabular}{@{}lllllllll@{}}
\toprule
Corr. Metric & Fairness Measurement & Logistic Regression & KLR\_Polynomial & LSIF & uLSIF $\alpha$ = 0.25 & uLSIF $\alpha$ = 0.5 & uLSIF $\alpha$ = 0.75 & uLSIF $\alpha$ = 1 \\ \hline

Independence & Logistic Regression & - & (0.03, -0.46*, 0.12) & (0.47*, 0.22, 0.04) & (0.60*, 0.13, -0.12) & (0.38*, 0.03, -0.32) & (0.21, 0.27, 0.27) & (-0.34*, 0.27, -0.01) \\
 & KLR\_Polynomial & (0.03, -0.46*, 0.12) & - & (0.04, -0.22, 0.35*) & (0.07, -0.3, 0.2) & (0.24, -0.04, 0.22) & (0.24, -0.27, 0.3) & (0.05, -0.27, 0.42*) \\
 & LSIF & (0.47*, 0.22, 0.04) & (0.04, -0.22, 0.35*) & - & (0.65*, 0.24, 0.06) & (0.33*, -0.01, -0.03) & (0.17, 0.35, 0.04) & (-0.24, 0.35, 0.06) \\
 & uLSIF $\alpha$ = 0.25 & (0.60*, 0.13, -0.12) & (0.07, -0.3, 0.2) & (0.65*, 0.24, 0.06) & - & (0.55*, 0.27, 0.4*) & (0.19, 0.72*, 0.3) & (-0.15, 0.72*, 0.24) \\
 & uLSIF $\alpha$ = 0.5 & (0.38*, 0.03, -0.32) & (0.24, -0.04, 0.22) & (0.33*, -0.01, -0.03) & (0.55*, 0.27, 0.4*) & - & (0.32*, 0.47*, 0.36) & (-0.0, 0.47*, 0.39) \\
 & uLSIF $\alpha$ = 0.75 & (0.21, 0.27, 0.27) & (0.24, -0.27, 0.3) & (0.17, 0.35, 0.04) & (0.19, 0.72*, 0.3) & (0.32*, 0.47*, 0.36) & - & (-0.11, 1.0*, 0.72*) \\
 & uLSIF $\alpha$ = 1 & (-0.34*, 0.27, -0.01) & (0.05, -0.27, 0.42*) & (-0.24, 0.35, 0.06) & (-0.15, 0.72*, 0.24) & (-0.0, 0.47*, 0.39) & (-0.11, 1.0*, 0.72*) & - \\ \hline

 Separation & Logistic Regression & - & (0.19, 0.54*, -0.01) & (0.67*, nan, 0.26) & (0.49*, nan, -0.05) & (0.28, nan, -0.26) & (0.26, nan, 0.27) & (0.03, nan, nan) \\
 & KLR\_Polynomial & (0.19, 0.54*, -0.01) & - & (0.04, nan, 0.17) & (0.4*, nan, 0.36*) & (0.35*, nan, 0.02) & (0.34*, nan, -0.27) & (0.07, nan, nan) \\
 & LSIF & (0.67*, nan, 0.26) & (0.04, nan, 0.17) & - & (0.28, nan, 0.13) & (0.22, nan, 0.05) & (0.11, nan, 0.28) & (-0.17, nan, nan) \\
 & uLSIF $\alpha$ = 0.25 & (0.49*, nan, -0.05) & (0.4*, nan, 0.36*) & (0.28, nan, 0.13) & - & (0.52*, nan, 0.22) & (0.34*, nan, 0.1) & (0.04, nan, nan) \\
 & uLSIF $\alpha$ = 0.5 & (0.28, nan, -0.26) & (0.35*, nan, 0.02) & (0.22, nan, 0.05) & (0.52*, nan, 0.22) & - & (0.19, nan, -0.18) & (-0.04, nan, nan) \\
 & uLSIF $\alpha$ = 0.75 & (0.26, nan, 0.27) & (0.34*, nan, -0.27) & (0.11, nan, 0.28) & (0.34*, nan, 0.1) & (0.19, nan, -0.18) & - & (0.12, nan, nan) \\
 & uLSIF $\alpha$ = 1 & (0.03, nan, nan) & (0.07, nan, nan) & (-0.17, nan, nan) & (0.04, nan, nan) & (-0.04, nan, nan) & (0.12, nan, nan) & - \\ \hline

Sufficiency & Logistic Regression & - & (-0.26, -0.42*, -0.19) & (-0.32*, -0.38*, -0.04) & (-0.53*, -0.09, -0.43*) & (-0.43*, -0.19, -0.44*) & (-0.34*, -0.27, -0.17) & (0.01, -0.27, -0.22) \\
 & KLR\_Polynomial & (-0.26, -0.42*, -0.19) & - & (0.45*, 0.12, -0.13) & (0.32*, 0.15, 0.16) & (0.32*, 0.23, -0.02) & (0.1, 0.22, 0.05) & (0.31*, 0.22, -0.22) \\
 & LSIF & (-0.32*, -0.38*, -0.04) & (0.45*, 0.12, -0.13) & - & (0.51*, 0.22, 0.05) & (0.38*, 0.15, -0.05) & (0.06, 0.35, -0.16) & (0.11, 0.35, -0.04) \\
 & uLSIF $\alpha$ = 0.25 & (-0.53*, -0.09, -0.43*) & (0.32*, 0.15, 0.16) & (0.51*, 0.22, 0.05) & - & (0.54*, 0.26, 0.41*) & (0.36*, 0.72*, -0.24) & (-0.06, 0.72*, -0.29) \\
 & uLSIF $\alpha$ = 0.5 & (-0.43*, -0.19, -0.44*) & (0.32*, 0.23, -0.02) & (0.38*, 0.15, -0.05) & (0.54*, 0.26, 0.41*) & - & (0.34*, 0.42*, 0.13) & (0.04, 0.42*, 0.2) \\
 & uLSIF $\alpha$ = 0.75 & (-0.34*, -0.27, -0.17) & (0.1, 0.22, 0.05) & (0.06, 0.35, -0.16) & (0.36*, 0.72*, -0.24) & (0.34*, 0.42*, 0.13) & - & (0.02, 1.0*, 0.72*) \\
 & uLSIF $\alpha$ = 1 & (0.01, -0.27, -0.22) & (0.31*, 0.22, -0.22) & (0.11, 0.35, -0.04) & (-0.06, 0.72*, -0.29) & (0.04, 0.42*, 0.2) & (0.02, 1.0*, 0.72*) & - \\ \hline
 
\end{tabular}%
}
\end{table*}

\begin{table*}[h]
\caption{Kendall correlations between pairs of methods approximating the Independence fairness metric at different mean intervals.
}
\label{tab:Kendall_intervall_updated_k}
\resizebox{\textwidth}{!}{%
\begin{tabular}{llrrrrr}
\hline
Mean Interval & Fairness Measurement & \multicolumn{1}{l}{Logistic Regression} & \multicolumn{1}{l}{Ridge} & Lasso & \multicolumn{1}{l}{KLR\_Gaussian} & \multicolumn{1}{l}{KLR\_Polynomial} \\ \hline
0 - 0.9 & Logistic Regression & - & 1.00 & 1.00 & 0.73 & 0.82 \\
 & Ridge & 1.00 & - & 1.00 & 0.73 & 0.82 \\
 & Lasso & 1.00 & 1.00 & - & 0.73 & 0.82 \\
 & KLR\_Gaussian & 0.73 & 0.73 & 0.73 & - & 0.55 \\
 & KLR\_Polynomial & 0.82 & 0.82 & 0.82 & 0.55 & - \\ \hline
1 - 1.9 & Logistic Regression & - & 1.00 & 1.00 & 0.91 & 0.73 \\
 & Ridge & 1.00 & - & 1.00 & 0.91 & 0.73 \\
 & Lasso & 1.00 & 1.00 & - & 0.91 & 0.73 \\
 & KLR\_Gaussian & 0.91 & 0.91 & 0.91 & - & 0.82 \\
 & KLR\_Polynomial & 0.73 & 0.73 & 0.73 & 0.82 & - \\ \hline
2 - 2.9 & Logistic Regression & - & 0.96 & 1.00 & 0.78 & 0.38 \\
 & Ridge & 0.96 & - & 0.96 & 0.73 & 0.33 \\
 & Lasso & 1.00 & 0.96 & - & 0.78 & 0.38 \\
 & KLR\_Gaussian & 0.78 & 0.73 & 0.78 & - & 0.42 \\
 & KLR\_Polynomial & 0.38 & 0.33 & 0.38 & 0.42 & - \\ \hline
3 - 3.9 & Logistic Regression & - & 0.78 & 0.96 & 0.56 & -0.02 \\
 & Ridge & 0.78 & - & 0.82 & 0.33 & 0.20 \\
 & Lasso & 0.96 & 0.82 & - & 0.51 & 0.02 \\
 & KLR\_Gaussian & 0.56 & 0.33 & 0.51 & - & -0.29 \\
 & KLR\_Polynomial & -0.02 & 0.20 & 0.02 & -0.29 & - \\ \hline
\end{tabular}%
}
\end{table*}

\begin{table*}[h]
\caption{Kendall correlations between pairs of methods approximating the Separation fairness metric at different mean intervals.}
\label{tab:2d_sep_Kendall_updated_k}
\resizebox{\textwidth}{!}{%
\begin{tabular}{lllllll}
\hline
Mean Interval & Fairness Measurement & Logistic Regression & Ridge & \multicolumn{1}{r}{Lasso} & KLR\_Gaussian & KLR\_Polynomial \\ \hline
0 - 0.9 & Logistic Regression & - & 0.99 & 0.9 & 0.72 & 0.81 \\
 & Ridge & 0.99 & - & 1.00 & 0.73 & 0.82 \\
 & Lasso & 0.99 & 1.00 & - & 0.73 & 0.82 \\
 & KLR\_Gaussian & 0.72 & 0.73 & 0.73 & - & 0.73 \\
 & KLR\_Polynomial & 0.81 & 0.82 & 0.82 & 0.73 & - \\ \hline
1 - 1.9 & Logistic Regression & - & 1.00 & 1.00 & 0.96 & 0.78 \\
 & Ridge & 1.00 & - & 1.00 & 0.96 & 0.78 \\
 & Lasso & 1.00 & 1.00 & - & 0.96 & 0.78 \\
 & KLR\_Gaussian & 0.96 & 0.96 & 0.96 & - & 0.73 \\
 & KLR\_Polynomial & 0.78 & 0.78 & 0.78 & 0.73 & - \\ \hline
2 - 2.9 & Logistic Regression & - & 0.96 & 1.00 & 0.87 & 0.33 \\
 & Ridge & 0.96 & - & 0.96 & 0.82 & 0.38 \\
 & Lasso & 1.00 & 0.96 & - & 0.87 & 0.33 \\
 & KLR\_Gaussian & 0.87 & 0.82 & 0.87 & - & 0.38 \\
 & KLR\_Polynomial & 0.33 & 0.38 & 0.33 & 0.38 & - \\ \hline
3 - 3.9 & Logistic Regression & - & 0.78 & 0.96 & 0.51 & 0.07 \\
 & Ridge & 0.78 & - & 0.82 & 0.29 & 0.29 \\
 & Lasso & 0.96 & 0.82 & - & 0.47 & 0.11 \\
 & KLR\_Gaussian & 0.51 & 0.29 & 0.47 & - & -0.24 \\
 & KLR\_Polynomial & 0.07 & 0.29 & 0.11 & -0.24 & - \\ \hline
\end{tabular}%
}
\end{table*}

\begin{table*}[]
\caption{Kendall correlations between actual and estimated Independence metric (with various core probabilistic classifiers) across various mean intervals.}
\label{tab:actuall_updated_k}
\centering
\resizebox{0.9\textwidth}{!}{%
\begin{tabular}{llrrrrr}
\hline
 Mean Interval & \multicolumn{1}{l}{Logistic Regression} & \multicolumn{1}{l}{Ridge} & \multicolumn{1}{l}{Lasso} & \multicolumn{1}{l}{KLR\_Gaussian} & \multicolumn{1}{l}{KLR\_Polynomial} \\ \hline

 0-0.9 & 0.94 & 0.94 & 0.94 & 0.67 & 0.85 \\
 1-1.9 & 1.00 & 1.00 & 1.00 & 0.91 & 0.73 \\
 2-2.9 & 0.91 & 0.96 & 0.91 & 0.78 & 0.38 \\
 3-3.9 & 0.87 & 0.64 & 0.82 & 0.69 & -0.16 \\
 \hline
\end{tabular}%
}
\end{table*}

\begin{table*}[]
\caption{Kendall correlations for the Independence fairness metric, computed using various measurement methods (relying on different classifiers), before (top) and after (bottom) thresholding their predicted probabilities.}
\label{tab:Kendall_threshold_updated_k}
\resizebox{\textwidth}{!}{%
\begin{tabular}{lllllll}
\hline
Corr. Metric &  & Logistic Regression & Ridge & Lasso & KLR\_Gaussian & KLR\_Polynomial \\ \hline
Kendall & Logistic Regression & - & 0.78 & 0.96 & 0.56 & -0.02 \\
(before thresholding) & Ridge & 0.78 & - & 0.82 & 0.33 & 0.20 \\
 & Lasso & 0.96 & 0.82 & - & 0.51 & 0.02 \\
 & KLR\_Gaussian & 0.56 & 0.33 & 0.51 & - & -0.29 \\
 & KLR\_Polynomial & -0.02 & 0.20 & 0.02 & -0.29 & - \\ \hline
Kendall & Logistic Regression & - & 1.00 & 1.00 & 0.78 & 0.91 \\
(after thresholding) & Ridge & 1.00 & - & 1.00 & 0.78 & 0.91 \\
 & Lasso & 1.00 & 1.00 & - & 0.78 & 0.91 \\
 & KLR\_Gaussian & 0.78 & 0.78 & 0.78 & - & 0.87 \\
 & KLR\_Polynomial & 0.91 & 0.91 & 0.91 & 0.87 & - \\ \hline
\end{tabular}%
}
\end{table*}

\begin{table*}[]
\caption{Kendall correlations for the Separation fairness metric, computed using various measurement methods (relying on different classifiers), before (top) and after (bottom) thresholding their predicted probabilities.}
\label{tab:Kendall_sep_thresh_updated_k}
\resizebox{\textwidth}{!}{%
\begin{tabular}{lllllll}
\hline
Corr. Metric &  & Logistic Regression & Ridge & Lasso & KLR\_Gaussian & KLR\_Polynomial \\ \hline
Kendall & Logistic Regression & - & 0.78 & 0.96 & 0.51 & 0.07 \\
(before thresholding) & Ridge & 0.78 & - & 0.82 & 0.29 & 0.29 \\
 & Lasso & 0.96 & 0.82 & - & 0.47 & 0.11 \\
 & KLR\_Gaussian & 0.51 & 0.29 & 0.47 & - & -0.24 \\
 & KLR\_Polynomial & 0.07 & 0.29 & 0.11 & -0.24 & - \\ \hline
Kendall & Logistic Regression & - & 1.00 & 1.00 & 0.73 & 0.56 \\
(after thresholding) & Ridge & 1.00 & - & 1.00 & 0.73 & 0.56 \\
 & Lasso & 1.00 & 1.00 & - & 0.73 & 0.56 \\
 & KLR\_Gaussian & 0.73 & 0.73 & 0.73 & - & 0.29 \\
 & KLR\_Polynomial & 0.56 & 0.56 & 0.56 & 0.29 & - \\ \hline
\end{tabular}%
}
\end{table*}

\end{document}